\documentclass[letterpaper, 10 pt, journal, twoside]{IEEEtran} 

\IEEEoverridecommandlockouts                              

\usepackage{graphicx}

\usepackage{amsmath,amssymb} 
\usepackage{color, soul}

\usepackage[lofdepth,lotdepth]{subfig}
\usepackage{float}
\usepackage{mathrsfs}
\usepackage{amsfonts}
\usepackage{algorithm}
\usepackage{algorithmic}
\usepackage{dashrule}

\usepackage{xr}
\usepackage{stix}

\title{HM$^4$: Hidden Markov Model with Memory Management for Visual Place Recognition}

\author{Anh-Dzung Doan, Yasir Latif, Tat-Jun Chin, and Ian Reid%
	\thanks{Manuscript received: July, 29, 2020; Revised October, 19, 2020; Accepted October, 22, 2020.}
	\thanks{This paper was recommended for publication by Editor Sven Behnke upon evaluation of the Associate Editor and Reviewers' comments.
		This work was supported by the ARC Centre of Excellence on Robotic Vision.} 
	\thanks{School of Computer Science, The University of Adelaide, Australia
		{\tt\footnotesize dung.doan@adelaide.edu.au}}%
	
	\thanks{Digital Object Identifier (DOI): see top of this page.}
}

\newcommand{\cQ}{\mathcal{Q}}
\newcommand{\cG}{\mathcal{G}}

\newcommand{\cC}{\mathcal{C}}
\newcommand{\cD}{\mathcal{D}}
\newcommand{\cP}{\mathcal{P}}

\newcommand{\bE}{\mathbf{E}}

\newcommand{\bd}{\mathbf{d}}

\newcommand{\bI}{\mathbf{I}}
\newcommand{\cI}{\mathcal{I}}

\newcommand{\bO}{\mathbf{O}}
\newcommand{\bo}{\mathbf{o}}

\newcommand{\bp}{\mathbf{p}}

\newcommand{\bq}{\mathbf{q}}

\newcommand{\bS}{\mathbf{S}}

\newcommand{\bNN}{\mathbf{N}}

\newcommand*{\mathcolor}{}
\def\mathcolor#1#{\mathcoloraux{#1}}
\newcommand*{\mathcoloraux}[3]{%
  \protect\leavevmode
  \begingroup
    \color#1{#2}#3%
  \endgroup
}

\definecolor{myorange}{RGB}{255,144,0}
\definecolor{myred}{RGB}{255,0,0}
\definecolor{mygreen}{RGB}{91,192,0}
\definecolor{myblue}{RGB}{0,206,255}
\definecolor{green}{RGB}{0,165,0}

\DeclareMathOperator*{\argmax}{arg\,max}
\DeclareMathOperator*{\argmin}{arg\,min}

\begin{document}

\maketitle

\begin{abstract}

Visual place recognition needs to be robust against appearance variability due to natural and man-made causes. Training data collection should thus be an ongoing process to allow continuous appearance changes to be recorded. However, this creates an unboundedly-growing database that poses time and memory scalability challenges for place recognition methods. To tackle the scalability issue for visual place recognition in autonomous driving, we develop a Hidden Markov Model approach with a two-tiered memory management. Our algorithm, dubbed HM$^4$, exploits temporal look-ahead to transfer promising candidate images between passive storage and active memory when needed. The inference process takes into account both promising images and a coarse representations of the full database. We show that this allows constant time and space inference for a fixed coverage area. 
The coarse representations can also be updated incrementally to absorb new data. To further reduce the memory requirements, we derive a compact image representation inspired by Locality Sensitive Hashing (LSH). Through experiments on real world data, we demonstrate the excellent scalability and accuracy of the approach under appearance changes and provide comparisons against state-of-the-art techniques.
\end{abstract}
\begin{IEEEkeywords}
	Localization, SLAM, Vision-Based Navigation.
\end{IEEEkeywords}

\section{Introduction} \label{sec:intro} 

\IEEEPARstart{V}{isual} Place Recognition (VPR) is the problem of localizing a camera based on visual input. Given a query image, a common localization approach is to retrieve an image from a geo-tagged database of images whose field of view (FOV) overlaps significantly with that of the query image. While the retrieval paradigm has received much attention~\cite{torii201524,arandjelovic2016netvlad}, current methods often fail when there are substantial appearance differences between the query and database.

It is crucial to realize that appearance changes occur continuously and indefinitely, due to factors that are natural (e.g.,~time of day, weather changes, vegetation growth) and man-made (e.g., construction works, update of billboards and facades). To build a data-driven VPR system that is robust against continuous appearance changes, a promising solution is to continuously accumulate data to refine the system~\cite{churchill2013experience,doan2019scalable,dymczyk2015gist}. While continuous data collection can be achieved via opportunistic or crowdsourced services (e.g., using taxi fleets), the ever-growing database demands a VPR algorithm scalable, i.e., the computational effort \emph{and} memory usage (to be distinguished from long-term storage) for training and inference must grow slowly with the database size.

In the context of autonomous driving, \cite{doan2019scalable}~presented a promising VPR solution based on Hidden Markov Model (HMM). The basic idea is to exploit temporal continuity in the trajectory of the car to guide image matching. A key innovation is the usage of a state space model (a topological map) and an observation model (which uses an image indexing structure) that can be iteratively updated to efficiently ``absorb" new information from newly appended images. This enables sublinear growth in \emph{inference time} w.r.t.~database size. \cite{doan2019scalable}~also showed that their updating procedure is much faster than the refinement process in end-to-end learning-based methods~\cite{brahmbhatt2018geometry}.

However, a fundamental weakness of~\cite{doan2019scalable} is the \emph{linear memory complexity} of the observation model w.r.t.~the database size. Specifically, while the image indexing structure \cite{muja2014scalable} that underpins the observation model can be queried and updated efficiently, the whole indexing structure must be loaded in the main memory to support querying---this is equivalent to storing \emph{all} database images in the main memory, which is infeasible in a lifelong operation.

A possible solution is to aggregate images (e.g.,~\cite{ni2009epitomic}) to remove redundancy and maintain a fixed-size database. However, image aggregation methods are imperfect and since the errors will be cumulated over time, the aggregated images will contain serious artefacts that affect localization accuracy.

A more direct solution is to progressively delete older images since they do not contain up-to-date appearance. However, devising an effective deletion scheme is nontrivial, since the images of different places could be refreshed at different rates (e.g., densely- versus sparsely-populated places), thus requiring careful tuning of place-dependent forgetting factors. More importantly, even if an effective deletion scheme can be built, the resulting database could still be too big to fit in memory---clearly a more fundamental treatment is required.

\begin{figure*}[ht]

	\centering
	\includegraphics[width=0.80\textwidth]{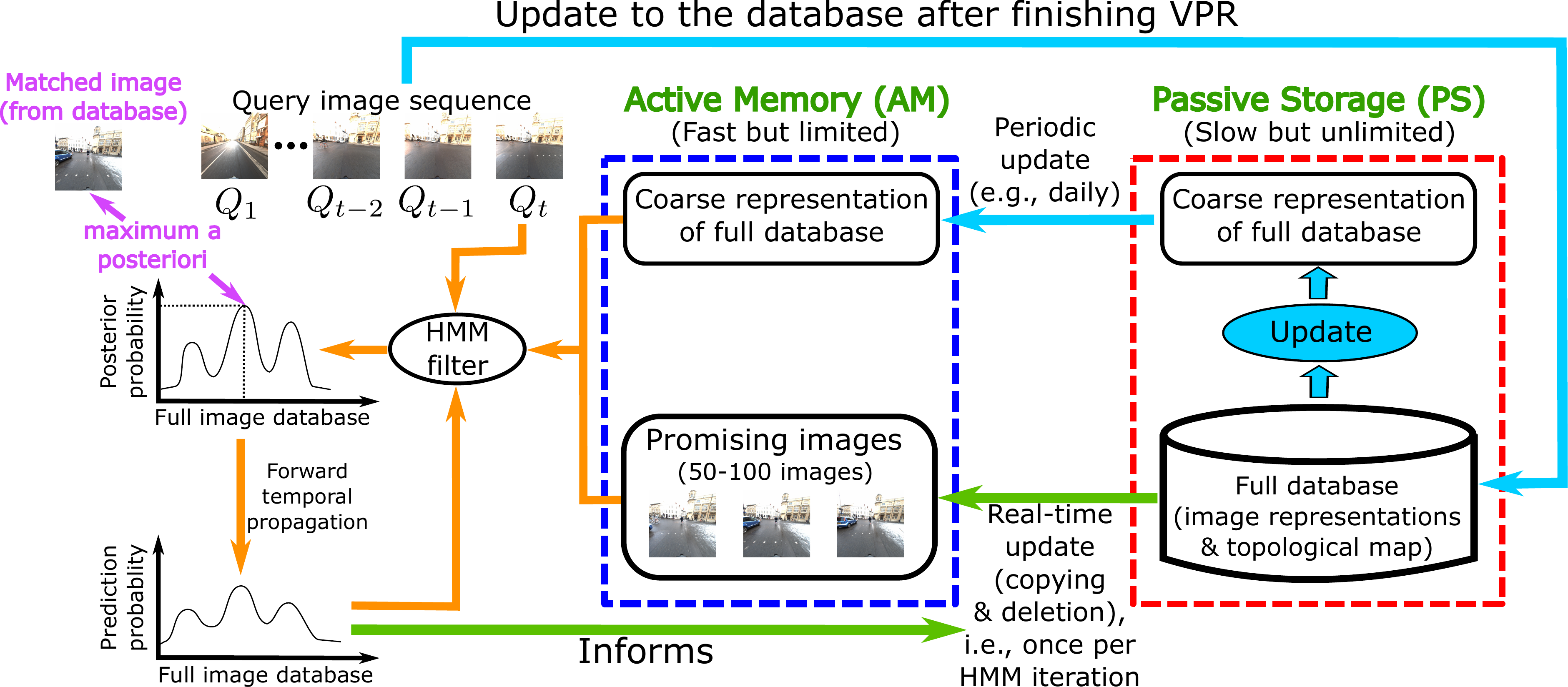}
	\caption{HM$^4$ system for lifelong VPR, where, $\mathcolor{myorange}{\pmb{\rightarrow}}$: online inference, $\mathcolor{mygreen}{\pmb{\rightarrow}}$: real-time update, and $\mathcolor{myblue}{\pmb{\rightarrow}}$: periodic update}
	\label{fig:twotier}
	\vspace{-0.5em}
\end{figure*}

\noindent\textbf{Contributions:}
We propose a novel HMM-based VPR system called HM$^4$ (HMM with Memory Management), which employs a two-tiered memory concept---active memory (AM) and passive storage (PS)---to allow scalable operation; see Fig.~\ref{fig:twotier}, AM corresponds to fast main memory whose size is limited, while PS represents slower long-term memory storing the full image database.

Given a query video $\cQ$, HM$^4$ compute the probability of matches between the current query image $Q_t \in \cQ$ and the full database. For scalability, temporal reasoning and image search are tightly coupled, and the probabilities are computed in a layered manner:
\begin{itemize}
	\item Forward temporal propagation is used to identify most promising database images to load to AM in real-time at each HMM iteration to replace the least promising ones. Typically, only a small number of promising images ($50$-$100$) are stored in AM at any time.
	
	\item A coarse representation of the full database in the form of feature space centroids and a topological submap (e.g., 7000 centroids and vertices for 17km traversal distance) are also loaded into AM, but these are updated at a lower frequency (e.g., daily) in PS to account for newly added image sequences to the database.
	
	\item The (posterior) probability of matches are recursively computed by the HMM filter, with transition and observation models that are evaluated exactly using the promising images, and approximately with bounded error using the coarse representations of the full database.
	
\end{itemize}
In effect, the data stored in AM is of constant size (provided the coverage area does not grow) thus ensuring HMM inference in constant time and memory, as well as robustness against appearance changes. Sec.~\ref{sec:hmm_scalable} will provide the details. 

In practice, we can implement AM on client side (e.g., self-driving cars) and PS on cloud whose storage can be seen as nearly infinite. The update operation is conducted via a stable and high-speed connection (e.g., 5G).

Our two-tiered memory concept for HMM-based VPR was inspired by RTAB-Map~\cite{labbe2013appearance} -- a loop closure detection for SLAM. RTAB-Map evaluates loop-closure hypotheses on images stored in AM (called ``working memory" in~\cite{labbe2013appearance}), which is a subset of the full image database judged to be the most promising by a heuristic. However, RTAB-Map does not maintain matching probabilities over the full database, thus, when none of the images in AM are good matches to the current query, the algorithm is lost and a hard reset is required. In contrast, HM$^4$ always maintains a probability distribution over the full database, thus allowing recognition of places not in the AM.

Furthermore, since the efficiency of VLAD~\cite{jegou2010aggregating} has been shown in place recognition, we derive its compact version called ``polytope VLAD" (polyVLAD), which greatly reduces the memory footprint of HM$^4$. Specifically, to avoid the burstiness problem, locally aggregated vectors are L2-normalized, which allows us to use a cross-polytope to partition the unit sphere. Inspired by LSH~\cite{andoni2015practical}, the compression scheme is conducted through random rotation matrices, which theoretically guarantees a small distance between two similar images (see Theorem 1 of~\cite{andoni2015practical}). Also, we propose to employ inverted index for effectively computing distances between polyVLAD vectors, leading to a significant improvement in terms of the inference time.

The experiments show HM$^4$ offers a scalability solution for VPR. Combined with polyVLAD, a lightweight system is obtained while localization accuracy is not influenced.

\section{Related work}

\subsection{Visual place recognition}
A popular approach in VPR is image retrieval from a geo-tagged database~\cite{cummins2008fab}. However, this approach is not robust towards large appearance changes. To address this issue, temporal continuity~\cite{milford2012seqslam} and appearance transfer~\cite{latif2018addressing} have been shown to be useful, but those did not consider lifelong operation where appearance changes occur continuously.

A number of works have addressed lifelong operation and/or continuous appearance changes by continuously accumulating training data~\cite{churchill2013experience,doan2019scalable,dymczyk2015gist}. Churchill et al.~\cite{churchill2013experience} rely on visual odometry (VO) for localization. Visual ``experiences" of places are maintained, which are trigged by localization failures. However, VO can be brittle, which adds complexity and failure points to the method. Doan et al.~\cite{doan2019scalable} address VPR using K-means tree-based image hypothesis generation loosely coupled with a HMM-based temporal fusion. A topological graph over image sequences is maintained whose nodes are equivalent to places in the scene. Query sequences are added to the database and the topological graph updated after place recognition hypothesis are formed. However, as alluded to above,~\cite{doan2019scalable} stores all image representations in memory, which is infeasible in a lifelong operation scenario. Latif et al.~\cite{latif2020sprint} exploits the sparsity in topological map for large-scale localization, but their approach do not fully address the memory scalability. Dymczyk et al.~\cite{dymczyk2015gist} maintain a scalable map by selecting a minimal number of landmarks. However, their sampling process is imperfect, which likely leads to discarding landmarks useful for localization.

\subsection{Compact image representation for place recognition}

Earlier methods used bag of words (BoW) representation \cite{galvez2012bags} on local features, while recent works \cite{torii201524, arandjelovic2016netvlad} showed VLAD \cite{jegou2010aggregating} is a more efficient technique. 

To obtain a lightweight system, quantization-based method~\cite{garg2020fast} is used to compress image representations to compact codes. Although this approach achieves a high accuracy in experiments, it does not offer the theoretical guarantees of similar images compressed to a same code---leading to a debate regarding its safety in long-term operation, in which appearance changes occur indefinitely.

By contrast, LSH~\cite{andoni2015practical} provided an established theory about nearest neighbor search. Recently, Lowry et al.~\cite{lowry2018lightweight} proposed to use LSH to compress VLAD vectors to binary codes. However, their method directly applied the hyperplane LSH to VLAD vectors without further exploring the characteristics of locally aggregated vectors on unit sphere.

\section{Background: HMM for VPR} \label{sec:hmm_overview}

Let $\cD = \{ I_i \}^{N}_{i=1}$ be a dataset of images (e.g., video frames of street recordings from multiple vehicles). Following~\cite{doan2019scalable}, we regard each $I_i$ as a ``place" and define a topological map $\cG$ over $\cD$, where each $I_i$ is a vertex of $\cG$. Two vertices are connected by an edge in $\cG$ if their FOVs overlap sufficiently (e.g., if they are temporally close in their source video, or if they are matched in previous iterations). Details on computing $\cG$ in our method will be provided in Sec.\ref{sec:hmm_scalable}.

Given a query video $\cQ = \{ Q_1, Q_2, \dots, Q_T \}$, our goal is to match each frame $Q_t \in \cQ$ with an image from $\cD$ that corresponds to the same place. To this end, we define an HMM $\{\bE, \bO, \pi\}$, where $\bE \in \mathbb{R}^{N\times N}$ is the transition matrix, $\bO \in \mathbb{R}^{N \times T}$ is the observation matrix, and $\pi \in \{0,1\}^N$, with $\sum_i \pi_i = 1$ is the initial matching probabilities (assumed to be uniform). The place associated to $Q_t$ is regarded as the random variable $s_t \in \{1,\dots,N\}$. Element $\bE(i,j)$ encodes the probability of moving from $I_i$ to $I_j$ between adjacent time steps $t-1$ and $t$
\begin{align}\label{eq:transitionmatrix}
	\bE(i, j) = P(s_t = j \mid s_{t-1} = i),
\end{align}
while element $\bO(i,t)$ encodes the probability of observing $Q_t$ given the place $I_i$
\begin{align}
	\bO(i, t) = P(Q_t \mid s_t = i).
\end{align}
Details on computing $\bE$ and $\bO$ in our method will be provided in Sec.~\ref{sec:hmm_scalable}.

%
%
Given the input sequence up to time $t$, i.e., $Q_{1:t} = \{ Q_1, Q_2, \dots, Q_t \}$, the aim is to calculate the posterior probabilities $P(s_t \mid Q_{1:t})$, for $s_t = 1,\dots,N$, which can be represented as a vector
\begin{align}
	\bp_t = \left[ \begin{matrix} P(1 \mid Q_{1:t}) & P(2 \mid Q_{1:t}) & \dots & P(N \mid Q_{1:t}) \end{matrix} \right].
\end{align}
The posterior probabilities are recursively computed as
\begin{equation}
\bp_t  = \eta \: \bo_t \circ \bE^T \bp_{t-1},
\label{eq:bayes_filter}
\end{equation}
where $\circ$ is the element-wise product, $\bo_t$ is the $t$-th column of $\bO$ and $\eta$ is a normalizing constant to ensure $\sum_i \bp_t(i) = 1$; for initialization, $\bp_0 = \pi$. The VPR decision at time $t$ is taken as the maximum \emph{a posteriori} result
\begin{equation}
s^*_t = \underset{i \in \{1,\dots,N\}}{\text{argmax}} \: \bp_t(i)
\label{eq:maxAP}
\end{equation}
For more details of the basic idea of HMM for VPR, see~\cite{doan2019scalable}.

From~\eqref{eq:bayes_filter}, it is clear that the time complexity of each HMM iteration is $O(Nr)$ ($r$ is the maximum number of non-zero values in each column of $\bE$), and memory complexity is $O(ND)$ ($D$ is dimensionality of image representation), where, $N$ increases unboundedly in lifelong operation.


\section{Achieving scalability with HM$^4$} \label{sec:hmm_scalable}

An overview of HM$^4$ was provided in Sec.~\ref{sec:intro} and Fig.~\ref{fig:twotier}. This section will describe the proposed VPR system in detail.

\subsection{Compact image representation}

Our system employs a new compact image representation called \emph{polytope VLAD}, which converts each image to a feature vector encoded by $8192$ bit. This provides a constant factor reduction in the computation and memory required. We postpone the description of polytope VLAD to Sec.~\ref{sec:polytope_vlad}; henceforth in this section, when we refer to an image, we mean specifically it is polytope VLAD vector.

\subsection{Coarse respresentation of full database} \label{sec:coarse_representation}

To avoid loading the full database $\cD = \{ I_i \}^{N}_{i=1}$ into AM, we extract a summary of the database in the form of a topological map and feature space centroids.

\subsubsection{Topological map} A topological map is built over $\cD$ to summarize the ``physical" connectivity of the images: in autonomous driving, the topological map reflects the road network of the coverage area. In HM$^4$, the topological map is equivalent to the transition matrix $\bE$~(Eq.~(\ref{eq:transitionmatrix})) of the HMM. Moreover, $\bE$ is built incrementally (sequence-by-sequence) as new videos are appended to $\cD$, using results of HMM inference. We will describe the initialization and updating of $\bE$ later in Sec.~\ref{sec:hmm_update}. For now,  assume $\bE$ is available.

\subsubsection{Feature space clustering}

\begin{figure}
	
	\centering	
	\includegraphics[width=0.3\textheight]{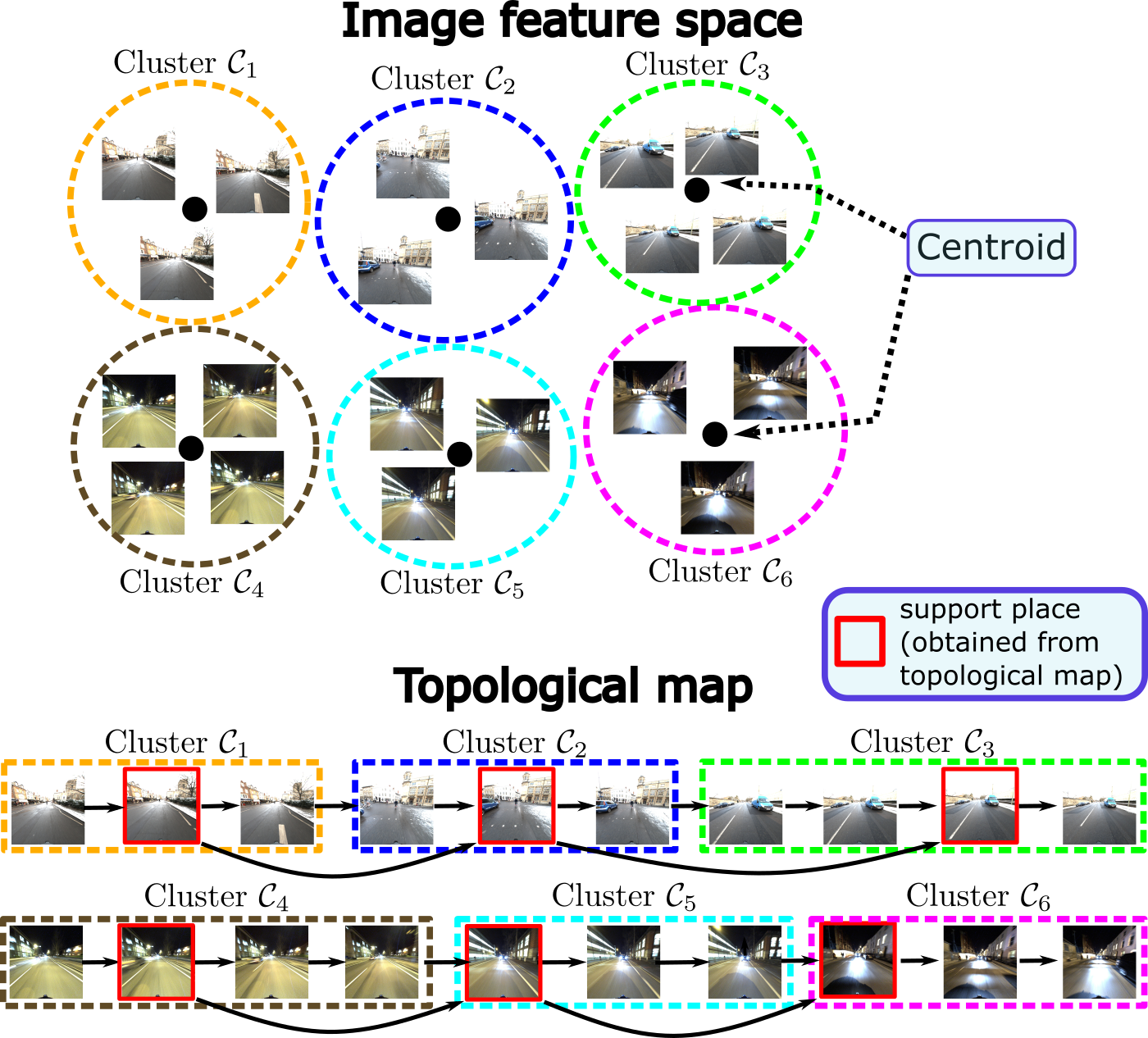}	
	\caption{Coarse representation of full database. The disjoint clusters are built on image feature space. Topological submap are created to coarsely represent the topological map.}
	\label{fig:coarse_representation}

\end{figure}

While $\bE$ summarizes the physical connectivity of $\cD$, the images are also clustered in the feature space to summarize their appearance. We partition $\cD$ (in feature space) into $K$ disjoint clusters $\{\cC_1, ..., \cC_{K}\}$, where each cluster $\cC_k$ is a subset of the images in $\cD$; for brevity, we also take
\begin{align}
	\cC_k \subset \{1,\dots,N\}
\end{align}
when we wish to refer to the indices of images that lie in each cluster. Each $\cC_k$ contains a centroid $B_k$ as a representative feature vector. We will be using the centroids to compute approximate distances between $Q_t$ and the images in $\cD$. If the chosen distance $d$ is a metric, it then holds that
\begin{align}\label{eq:triang}
	\forall I_i \in \cC_k, \;\; |d(Q_t, I_i) - d(Q_t, B_k)| < d(I_i, B_k).
\end{align}
We then take $d(Q_t,I_i) \approx d(Q_t,B_k)$ for all $I_i \in \cC_k$. To minimize the approximation error, we should find clusters $\{\cC_k\}^{K}_{k=1}$ and centroids $\{ B_k \}_{k=1}^{K}$ that minimize
\begin{align}\label{eq:densevlad_all_cluster_cost}
	\textstyle\frac{1}{N} \sum_{k=1}^{K} \sum_{I_i \in \cC_k}d(I_i, B_k).
\end{align}
For continuous feature spaces (e.g., $I_i$ is a NetVLAD~\cite{arandjelovic2016netvlad} vector) where $d$ is the Euclidean distance, K-means algorithm can be used to perform the clustering. The proposed polytope VLAD, however, yields discrete feature vectors; we will discuss the appropriate metric in Sec.~\ref{sec:polytope_vlad}.

Also, similar to the construction of $\bE$, the feature space clusters are computed incrementally as new videos are appended to $\cD$, which we will describe in Sec.~\ref{sec:hmm_update}.


\subsubsection{Topological submap}

Note that the centroids $\{ B_k \}_{k=1}^K$ do not generally correspond to feature vectors of actual images. To associate an image to each $B_k$, we seek the image in $\cC_k$ that has the highest degree in $\bE$, i.e.,
\begin{equation}
p_k = \textstyle\argmax_{i \in \cC_k} \sum_{j} \mathbb{I}(\bE(i,j) > 0).
\label{eq:support_place}
\end{equation}
We call $I_{p_k}$ the \emph{support place} of $\cC_k$. Given the clusters and associated support places, we construct the topological submap of $\cD$ as
\begin{align}
	\bE_{sm} = \left[ \begin{matrix} \bE(:,p_1) & \bE(:,p_2) & \dots & \bE(:,p_K)\end{matrix} \right].
\end{align}
In words, $\bE_{sm}$ are the columns of $\bE$ corresponding to the support places. See Fig.~\ref{fig:coarse_representation} for a high-level idea of the coarse representation of $\cD$.

\subsection{Two-tiered HMM inference}\label{sec:hmm_memory}\
\vspace{-1em}

\begin{figure*}
	
	\centering
	
	\includegraphics[width=0.75\textwidth]{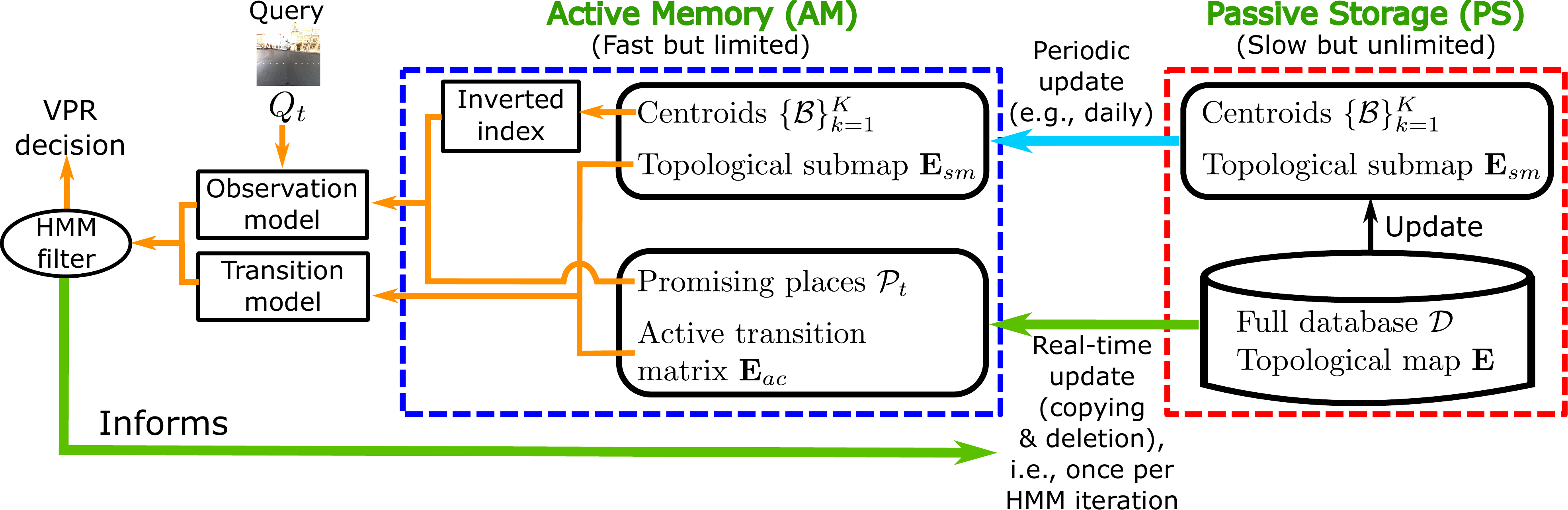}
	
	\caption{Data available in AM to perform VPR inference, where, $\mathcolor{myorange}{\pmb{\rightarrow}}$: online inference, $\mathcolor{mygreen}{\pmb{\rightarrow}}$: real-time update, and $\mathcolor{myblue}{\pmb{\rightarrow}}$: periodic update. Note that ``Inverted index", which is used for polytope VLAD, will be described in Sec.~\ref{sec:polytope_vlad}. Before Sec.~\ref{sec:polytope_vlad}, we can ignore this component, and assume centroids are the input of observation model.}
	\label{fig:hm4_data_needed}
	\vspace{-0.5em}
\end{figure*}

Our aim in this subsection is to perform HMM inference to obtain the posterior $\bp_t$ (over the full database) for the current time $t$. Fig.~\ref{fig:hm4_data_needed} illustrates the data available to perform the inference in HM$^4$. Specifically, only the coarse representations of the full database---specifically, $\{ B_k \}^{K}_{k=1}$ and $\bE_{sm}$---are stored in AM, whereas the full database $\cD$ and topological map $\bE$ are stored in PS.

Noting that HMM inference is a recursive process, using the posterior probability $\bp_{t-1}$ from the previous iteration, we identify a set of \emph{promising places}:
\begin{equation}
\mathcal{P}_t = \cP_t^{(1)} \cup \cP_t^{(2)},
\label{eq:promising_set}
\end{equation}
where, $\mathcal{P}^{(1)}_t = \{ i \in \{ 1,\dots,N\} \mid \bp_{t-1}(i) \ge \zeta \}$, $\mathcal{P}^{(2)}_t = \{ j \in \{ 1,\dots,N\} \mid \bE(i,j) > 0, i \in \cP^{(1)}_t \},$ 
and $\zeta \in [0,1]$ is a preselected threshold (see Sec.~\ref{sec:hmm_overall_algo} on setting $\zeta$). Intuitively, $\mathcal{P}_t$ consists of the set of places that $Q_t$ likely corresponds to according to HMM propagation (before considering the appearance of $Q_t$), and the images that are adjacent to the likely places according to the topological map. It is vital to highlight that $\cP_t$ is a tiny subset (e.g., $|\cP_t| \le 100$) of the full database $\cD$.

We then obtain the \emph{active} transition matrix as
\begin{equation}
\bE_{ac} = \left[ \begin{matrix} \bE(:,\cP_{t,1}) & \bE(:,\cP_{t,2}) & \dots & \bE(:,\cP_{t,M})  \end{matrix} \right] \in \mathbb{R}^{N \times M},
\label{eq:active_transition_matrix}
\end{equation}
where $\cP_{t,m}$ is the $m$-th item of $\cP_t$, and $M = |\cP_t|$. In words, $\bE_{ac}$ are the columns of $\bE$ corresponding to the promising places. Since $\bE$ resides in PS, $\bE_{ac}$ is transferred to the AM at every HMM iteration. Also, while the size of $\bE_{ac}$ depends on $N$, the matrix is very sparse, thus the transfer is cheap.

Given the current query image $Q_t$, our HMM observation model is based on computing the observation likelihood
\begin{equation}
L(Q_t,I_i) = \exp\left(-\tfrac{d(Q_t,I_i)}{\sigma}\right),
\label{eq:obs_likelihood}
\end{equation}
where $I_i$ is an arbitrary database image, $d$ is the feature space metric, and $\sigma$ is a bandwidth parameter (see Sec.~\ref{sec:hmm_overall_algo} for its setting). To avoid computing the likelihood over the full database, we define the \emph{active} observation vector
\begin{equation}
\small
\bo_{t,ac} = \left[ \begin{matrix} L(Q_t,I_{\cP_{t,1}}), L(Q_t,I_{\cP_{t,2}}), \dots , L(Q_t,I_{\cP_{t,M}})  \end{matrix} \right]^T  \in \mathbb{R}^{M},
\label{eq:active_obs_vector}
\end{equation}
which requires only the promising images, and the \emph{background} observation vector
\begin{equation}
\bo_{t,bg} = \left[ \begin{matrix} L(Q_t,B_1) , L(Q_t,B_2) , \dots , L(Q_t,B_K)  \end{matrix} \right]^T \in \mathbb{R}^K,
\label{eq:background_obs_vector}
\end{equation}
which uses the feature space centroids of the full image database. As in the case of obtaining the active transition matrix, the feature vectors of the promising places will be transferred to AM at each HMM iteration.

We can now perform the inference: we modify the standard HMM update~\eqref{eq:bayes_filter} to use only the available information in AM to obtain the matching confidence:

\begin{itemize}
	\setlength{\parskip}{0pt}
	\setlength{\itemsep}{0pt plus 1pt}
	\item Matching confidence in promising places $\cP_t$: 
	\begin{equation}
	\bq_t(\cP_{t,m}) = \bo_{t,ac}(m).\bE_{ac}(:,m)^T.\bp_{t-1},
	\label{eq:matching_confidence_promising_places}
	\end{equation}
	
	\item Matching confidence in other places computed from coarse representation:
	\begin{equation}
	\bq_t(i) = \bo_{t,bg}(k).\bE_{sm}(:,k)^T.\bp_{t-1}, \:\: \text{if} \,\, i \notin \cP_t \:\: \& \:\: i \in \cC_k,
	\label{eq:matching_confidence_background_places}
	\end{equation}
\end{itemize}
The posterior $\bp_t$ is then obtained by normalizing:
\begin{equation}
\bp_t = \displaystyle\frac{\bq_t}{\sum \bq_t},
\label{eq:belief}
\end{equation}

\noindent The VPR decision is finally made by (Eq. (\ref{eq:maxAP})).

\subsection{Updating database representation}\label{sec:hmm_update}
\begin{figure}[h]
	
	\centering
	\mbox
	{
		\subfloat[][]
		{
			\includegraphics[width=0.3\textheight]{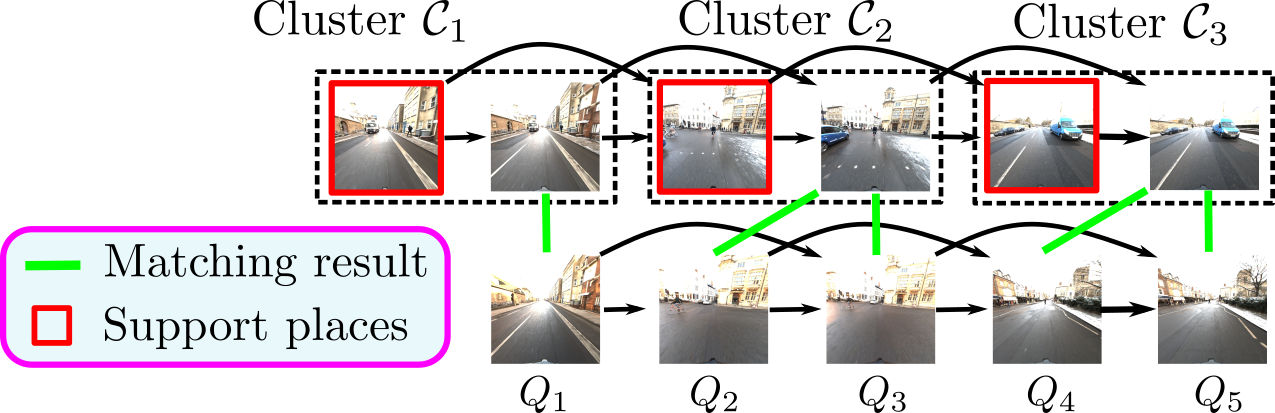}
			\label{fig:hmm_matching}
		}
		
	}
	\mbox
	{
		\subfloat[][]
		{
			\includegraphics[width=0.32\textheight]{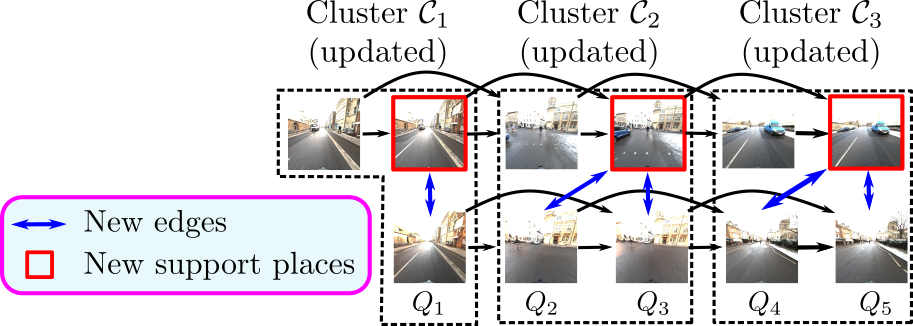}
			\label{fig:hmm_update}
		}
	}
	
	\caption{\textbf{(a)} Matching results for $\cQ = \{Q_i\}_{i=1}^5$. \textbf{(b)} Updating video $\cQ$ to the database.}
	\label{fig:hm4_matching_update}
	
\end{figure}

Given a query video $\cQ = \{Q_t\}_{t=1}^T$ localized by VPR.

\subsubsection{Update topological map} We create its topological map 
\begin{equation}
\small
\bE^{\cQ}(i,j) = \begin{cases}
\alpha \: \text{exp}\frac{(j - i)^2}{\delta^2} & 0 \le j - i \le V_{\text{max}}\\
0 & \text{otherwise} 
\end{cases}
\label{eq:init_transition}
\end{equation}
where, $\alpha$ is normalization constant to ensure the summation of every row of $\bE^{\cQ}$ equals to one, $V_{\text{max}}$ is the maximum velocity of the vehicle, and $\delta$ is preselected scale value (see Sec.~\ref{sec:hmm_overall_algo} for their settings).
After that, the topological map $\bE$ is expanded in the diagonal direction
\begin{align}
	\small
	\bE = \begin{bmatrix}\bE & 0_{N \times T} \\ 0_{T \times N} & \bE^{\cQ} \end{bmatrix}.
\end{align}
We create new edges for every localized query $Q_t \in \cQ$ with its matched place $i$:
\begin{align}
\begin{array}{c}
    \bE(t+N, i) = \bE(t+N, t+N) \\
    \bE(i, t+N) = \bE(i, i).
\end{array}
\end{align}
$\bE$ is then normalized to ensure the summation of each row equals to 1.

\subsubsection{Update coarse representation} 
For each $Q_t \in \cQ$, we add it to its matched cluster, then compute new centroids:
\begin{equation}
B_k = \underset{B_k}{\argmin} \textstyle\sum_{I_i \in \cC_k}d(I_i, B_k)
\label{eq:updating_centroid}
\end{equation}
For continuous feature space, where $d$ is Euclidean distance, $B_k = \textstyle\frac{1}{|\cC_k|}\textstyle\sum_{I_i \in \cC_k}I_i$. For polytope VLAD, Sec.~\ref{sec:polytope_vlad} will discuss how to compute $B_k$.

With new centroids and new topological map $\bE$ expanded by $\bE^{\cQ}$, we build a new topological submap $\bE_{sm}$ as described in Sec.~\ref{sec:coarse_representation}.

Fig.~\ref{fig:hm4_matching_update} shows the high-level idea of the updating process. Note that this updating process also allows us to start with database $\cD$ containing a single video first, then incrementally build topological map and coarse representation (sequence by sequence) using HMM inference.

\subsection{Overall algorithm} \label{sec:hmm_overall_algo}

The proposed algorithm for VPR is presented in Algorithm \ref{algo:pr_hmm}, and its high-level idea is shown in Fig~\ref{fig:twotier}. At the first frame ($t=1$), $\bp_0$ is uniformly distributed, hence we only use  $\bo_{t,bg}$ for computing belief $\bp_1$. In our experiment, we set $V_{max} = 10$, $\delta = 3$, $\sigma = 0.03$, and $\zeta = 0.00015$.

\begin{algorithm}[t] \centering
	\small
	\begin{algorithmic}[1]
		\REQUIRE Unlimited query videos $\{\cQ_i\}_{i=1}^{\infty}$, topological map $\bE$, database $\cD$, parameters $V_{max}$, $\delta$, $\sigma$, $\zeta$
		\STATE Build clusters $\{\cC\}_{k=1}^K$, centroids $\{B\}_{k=1}^K$, and topological submap $\bE_{sm}$ (Sec. \ref{sec:coarse_representation})
		\STATE Copy centroids $\{B\}_{k=1}^K$, and topological submap $\bE_{sm}$ from PS to AM
		\FOR {\text{each} $\cQ \in \{\cQ_i\}_{i=1}^{\infty}$}
		\STATE Initialize $\cP_0 = \emptyset$
		\FOR {\text{each frame} $Q_t \in \cQ$}
		
		\IF{$t = 1$}
		\STATE Compute $\bo_{t,bg}$ using Eq. (\ref{eq:background_obs_vector})
		\STATE Compute matching confidence $\bq_1$ using Eq. (\ref{eq:matching_confidence_background_places})
		
		\ELSE
		\STATE Seek $\cP_t$ using Eq. (\ref{eq:promising_set})
		\STATE Build $\bE_{ac}$ using Eq. (\ref{eq:active_transition_matrix})
		\STATE Copy images $\{I_i \, | \, i \in \cP_t, \, i \notin \cP_{t-1}\}$ from PS to AM
		\STATE Delete images $\{I_i \, | \, i \notin \cP_t, \, i \in \cP_{t-1}\}$ in AM
		\STATE Compute $\bo_{t,ac}$ using Eq. (\ref{eq:active_obs_vector}), and $\bo_{t,bg}$ using Eq. (\ref{eq:background_obs_vector})
		\STATE Compute matching confidence $\bq_t$ using (\ref{eq:matching_confidence_promising_places}) and (\ref{eq:matching_confidence_background_places})
		
		\ENDIF
		\STATE Compute belief $\bp_t$ using Eq. (\ref{eq:belief})
		\STATE Find matched place/image $i^*$ using Eq. (\ref{eq:maxAP})
		
		\ENDFOR
		\STATE Store localized $\cQ$ in PS for updating.
		\IF {(On map update)} \label{algo:on_map_update}
		\STATE Update topological map $\bE$, clusters $\{\cC\}_{k=1}^K$, centroids $\{B\}_{k=1}^K$, and topological submap $\bE_{sm}$ (Sec.~\ref{sec:hmm_update})
		\STATE Copy new centroids $\{B\}_{k=1}^K$, new topological submap $\bE_{sm}$ from PS to AM
		\ENDIF
		\ENDFOR

	\end{algorithmic}
	\caption{VPR with HM$^4$}
	\label{algo:pr_hmm}
	
\end{algorithm}

\noindent\textbf{Complexity analysis:} \label{sec:complexity_analysis_system} In standard HMM implementation, at each HMM iteration (each $Q_t$), the memory complexity is $O(ND)$, and time complexity of computing $\bo_t$ is $O(ND)$, that of computing $\bp_t$ is $O(Nr)$ ($r$ is the numbers of non-zero values in each column of $\bE$, which can be seen as a constant). In the context of sequentially updating new sequences, $N$ grows \emph{linearly and unboundedly}. 

By contrast, our VPR algorithm has $O\begin{pmatrix}(K + |\cP_t|)D \end{pmatrix}$ memory complexity, $O\begin{pmatrix}(K + |\cP_t|)D \end{pmatrix}$ time complexity of computing $\bo_t$ ($\bo_{t,bg}$ and  $\bo_{t,ac}$), and $O\begin{pmatrix}(K + |\cP_t|)r \end{pmatrix}$ time complexity of computing $\bp_t$ (note $K + |\cP_t| \ll N$).  We experimentally show $K + |\cP_t|$ remains almost constant when updating new sequences. In Sec.~\ref{sec:polytope_vlad}, with polytope VLAD, the time complexity of computing $\bo_t$ can be slashed by a constant factor, i.e.,  $O\begin{pmatrix}\frac{KD}{\text{const}} + |\cP_t|D \end{pmatrix}$

\section{Polytope VLAD} \label{sec:polytope_vlad}

This section will describe polytope VLAD in details (Sec.~\ref{sec:encode_polyVLAD}), and the usage of inverted index for efficiently computing observation model (Sec.~\ref{sec:inverted_index}).

\subsection{Background: VLAD}
VLAD \cite{jegou2010aggregating} is a feature aggregations scheme that
given a set of local image features $\{f_i\}_{i=1}^{\bNN}$ (either hand-crafted \cite{torii201524} or deep learning \cite{gong2014multi} features), it assigns each local feature $f_i$ to the closest cluster $c_l$ of a vocabulary of size $L$, and accumulates residuals via:
$
x_l = \sum_{f_i \in \text{NN}(c_l)} f_i - c_l
$

To avoid burstiness problem, before concatenating all $x_l$ to form a VLAD vector, $x_l$ is L2-normalized \cite{arandjelovic2013all}, thus each $x_l$ is located on the unit sphere $S^{\bd-1}$. Adapted the idea of LSH \cite{andoni2015practical}, we derive the ``polytope VLAD'' (polyVLAD) to compress vectors $x_l$ into compact codes via a cross-polytope.

\begin{figure}
	
	\centering
	\mbox
	{
		\subfloat[][polyVLAD idea]
		{
			\includegraphics[width=0.16\textwidth]{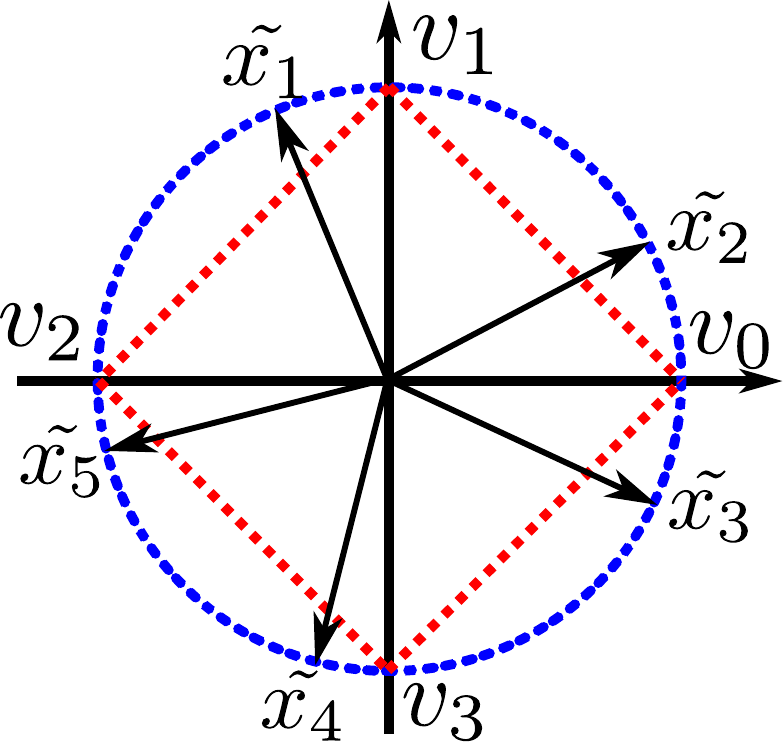}
			\label{fig:polyVLAD}
		}
		\subfloat[][An example of inverted index]
		{
			\includegraphics[width=0.28\textwidth]{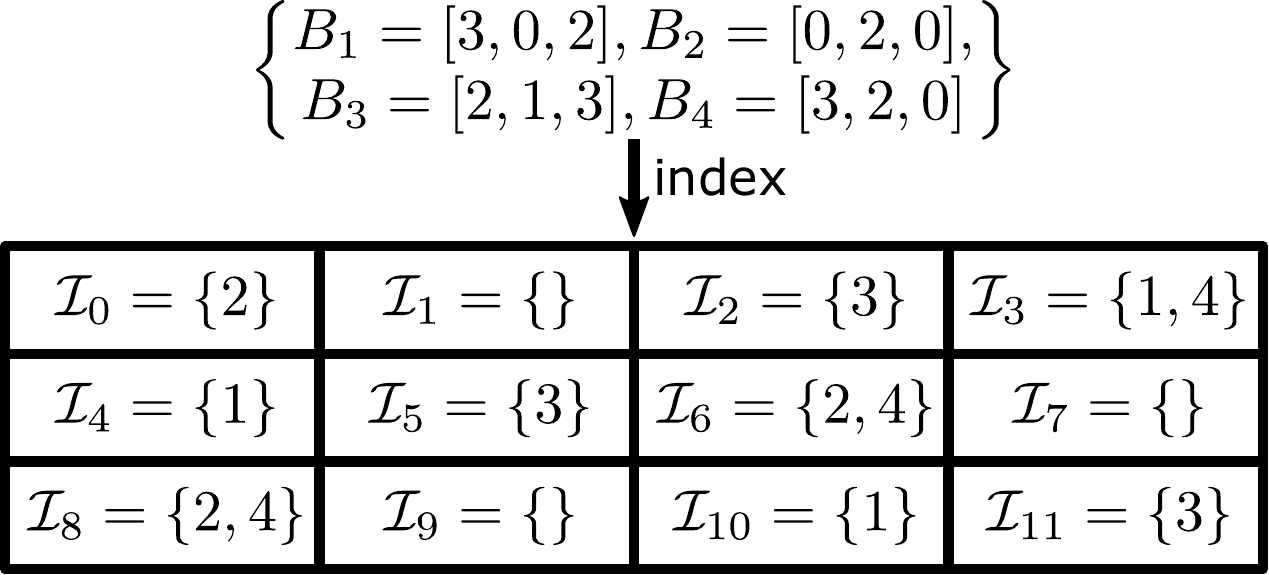}
			\label{fig:inverted_index}
		}
	}
	
	\caption{\textbf{(a)} A polytope (dashed red) with vertices $v_0$, $v_1$, $v_2$, $v_3$ is used to partition a unit sphere (dashed blue). \textbf{(b)} The size of inverted index is $D\times2\bd$ ($D= 3$, $\bd =2$). Centroid $B_1$ is indexed to $\cI_3$, $\cI_4$ and $\cI_{10}$ using equation (\ref{eq:find_inverted_index_bucket}).} 

\end{figure}

\subsection{Polytope VLAD} \label{sec:encode_polyVLAD}

Given a set of vectors $X = [x_1, ..., x_L]$ on a unit sphere $S^{\bd-1}$,
we employ a cross-polytope to partition the unit sphere $S^{\bd-1}$. 
A cross-polytope is defined by a
set of vertices: $V = \{v_i \: | \: i = 0, ..., 2\bd-1 \}$ that are all the permutations of $(\pm1, 0,...,0)$, where $v_i$ is a $\bd$-dimensional vector. Let $R \in \mathbb{R}^{\bd \times \bd}$ be a random rotation matrix. We firstly rotate $X$ by $R$: $\tilde{X} = RX = [\tilde{x_1}, ..., \tilde{x_L}]$, 
then each rotated vector $\tilde{x_l}$ is encoded by the nearest vertex of the polytope $V$. The idea is illustrated in Fig.~\ref{fig:polyVLAD} and the encoding function is:

\begin{equation}
h_l = \underset{i}{\text{argmax}}(\tilde{x}^T_l.v_i) \: , \forall v_i \in V.
\label{eq:poly_encode_1}
\end{equation}
Rearranging all the vertex vectors $v_i$ in a matrix:
$
V_{\bd \times 2\bd} =
\left[
\begin{matrix}
\bI_{\bd \times \bd} & -\bI_{\bd \times \bd}
\end{matrix}
\right]
$
where, $\bI_{\bd \times \bd}$ is identity matrix of size $\bd \times \bd$, the product of $V^T.\tilde{x}_l$ is vector
\begin{align}
	u_l = & [\tilde{x}^{[1]}_l, \dots, \tilde{x}^{[\bd]}_l, -\tilde{x}^{[1]}_l, \dots, -\tilde{x}^{[\bd]}_l]^T 
	= [\tilde{x}_l^T, -\tilde{x}_l^T]^T
\end{align}
Due to the symmetric characteristic of $u_l$, we rewrite the encoding function (\ref{eq:poly_encode_1}):

\begin{equation}
h_l = 
\begin{cases} 
m^* - 1 & \tilde{x}^{[m^*]}_l \ge 0 \\
m^* + \bd -1 & \tilde{x}^{[m^*]}_l < 0 
\end{cases}
\label{eq:polyVLAD_enc}
\end{equation}
\noindent where $m^* = \underset{m\in\{1,...,d\}}{\argmax} (|\tilde{x}^{[m]}_l|)$ and $h_l$ is the index of the closest vertex to $\tilde{x}_l$. With $M$ different rotation matrices, $M$ polyVLAD vectors are generated and then concatenated to obtain single polyVLAD vector $I$. The number of bits required to encode vector $I$ is: $(\text{log}_2\bd + 1).L.M$

\subsubsection{Clustering polyVLAD} As polyVLAD vectors are in discrete space, Jaccard distance (also metric) \cite{kosub2019note} is used. To build clusters $\{\cC\}_{k=1}^K$ for coarse representation of HM$^4$, we use K-modes algorithm \cite{huang1998extensions} to minimize (\ref{eq:densevlad_all_cluster_cost}). In updating process (Sec.~\ref{sec:hmm_update}), new centroids (equation (\ref{eq:updating_centroid})) are computed using Theorem 1 of \cite{huang1998extensions}.

\subsection{Inverted index for efficient distance computation} \label{sec:inverted_index}

In algorithm \ref{algo:pr_hmm}, computing $\bo_{t,bg}$ (equation (\ref{eq:background_obs_vector})) is most costly, as numbers of clusters are much larger than numbers of promising images ($K \gg |\cP_t|$). Therefore, inverted index is used to index centroids $\{B_k\}_{k=1}^K$ (see Fig.~\ref{fig:hm4_data_needed}). 

Given a set of centroids $\{B_k\}_{k=1}^K$ and a query $Q_t = \begin{bmatrix}q^{[1]},..., q^{[D]}\end{bmatrix}$ (note: $q^{[m]}$ is a scalar) represented by polyVLAD, we aim to calculate Jaccard distance from $Q_t$ to every centroid $B_k$. We use an inverted index: $\{\cI_0, ..., \cI_{W-1}\}$, whose size is $W = 2\bd L M = 2\bd D$. 
Each $B_k = \begin{bmatrix}b^{[1]}_k,..., b^{[D]}_k\end{bmatrix}$ (note: $b_k^{[m]}$ is a scalar) is indexed:
\begin{equation}
\forall m = \{1,...,D\}, \: \cI_{2\bd(m-1) + b^{[m]}_k} = \cI_{2\bd(m-1) + b^{[m]}_k} \cup k
\label{eq:find_inverted_index_bucket}
\end{equation}
Fig. \ref{fig:inverted_index} is an example of inverted index. In the online inference, given a query: $Q_t = \begin{bmatrix}q^{[1]},..., q^{[D]}\end{bmatrix}$. We first calculate similarity score between $Q_t$ to every $B_k$:

\begin{itemize}
	
	\item Initialize similarity score $\bS(Q_t,B_k) = 0$ for every $B_k$
	\item For each $m \in \{1, ..., D\}$: $\bS(Q_t,B_k) \leftarrow \bS(Q_t,B_k) + 1$, for every $k \in \cI_{2\bd(m-1) + q^{[m]}}$
	
\end{itemize} 
Jaccard distances are then computed: $d(Q, B_k) = 1 - \textstyle\frac{\bS(Q, B_k)}{D}$ for every $B_k$. Finally, $\bo_{t,bg}$ is calculated from \eqref{eq:obs_likelihood} and \eqref{eq:background_obs_vector}.

\noindent\textbf{Complexity analysis:}
Assume that polytope vertex indices from $\{0, ..., 2\bd-1\}$ are distributed uniformly in each dimension $m$ of $\{B_k\}_{k=1}^K$. The complexity of computing distance from $Q_t$ to every $B_k$ is $O(\textstyle\frac{KD}{2\bd})$, i.e., $2\bd$ times faster than linear scan $O(KD)$. For some common local features, $\bd$ is usually larger than 100 (e.g., $\bd=128$ for SIFT feature). 

\vspace{-0.5em}
\section{Experiments}

\noindent\textbf{Datasets:} 2 datasets are used:
\begin{itemize}
    \item Oxford RobotCar~\cite{maddern20171}: 4 sequences (``26/06/2014, 09:24:58", ``26/06/2014, 08:53:56", ``23/06/ 2014, 15:41:25", and ``23/06/2014, 15:36:04") are used and briefly referred to S1, S2, S3 and S4. The traversal distance in each sequence is about 1km.
    \item St Lucia~\cite{glover2010fab}: 5 sequences (``10/09/2009, 08:45", ``10/09/2009,10:00",  ``19/08/2009, 08:45", ``21/08/2009, 10:10", and ``21/08/2009, 12:10") are used and briefly referred to A1, A2, A3, A4 and A5. The traversal distance in each sequence is about 17km.
\end{itemize}

\noindent\textbf{Implementation:}  We densely extract SIFT features at 4 scales with 16, 24, 32, 40-pixel region width, over a grid of 2-pixel spacing, and use vocabulary of size $128$ for polyVLAD. $8$ rotation matrices are sampled, resulting in each image being encoded by $8192$ bits. We set number of clusters $K=700$ and $7000$ respectively for Oxford RobotCar and St Lucia. Other parameters are set as in Sec.~\ref{sec:hmm_overall_algo}. All experiments are conducted on a computer with Intel Core i7@3.4GHz (8 cores) and 16GB RAM, where we simulate PS as the hard disk and AM as the RAM.

\noindent\textbf{Evaluation:} Our method (denoted as \texttt{polyVLAD+HM$^{4}$}) is compared against baseline, i.e., \texttt{SPR} \cite{doan2019scalable}, DenseVLAD \cite{torii201524} and NetVLAD \cite{arandjelovic2016netvlad} incorporated to HMM framework (denoted as \texttt{DenseVLAD+HMM} and \texttt{NetVLAD+HMM}). A query image is regarded correctly localized if the distance from the matched place to the ground truth position is less than a threshold varied from 1m to 25m. We also measure memory (RAM) needed to store topological map and database for inference, as well as the inference time.

To simulate the scenario of continuously accumulating image data. In Oxford Robotcar, S1 forms the initial database; S2, S3 and S4 are sequentially used as the query sequences. In St Lucia, A1 is selected as initial database; A2, A3, A4 and A5 are sequentially used as the query sequences. After each query sequence finishes, we update to the database based on VPR decision. For \texttt{DenseVLAD+HMM} and \texttt{NetVLAD+HMM}, we update the topological map as described in Sec.~\ref{sec:hmm_update}, and append image representations to database.

\subsection{Ablation study}\label{sec:exp_ablation}
\begin{figure}
	\centering
	\mbox
	{
	    \subfloat[][]
		{
		    \includegraphics[width=0.44\textwidth]{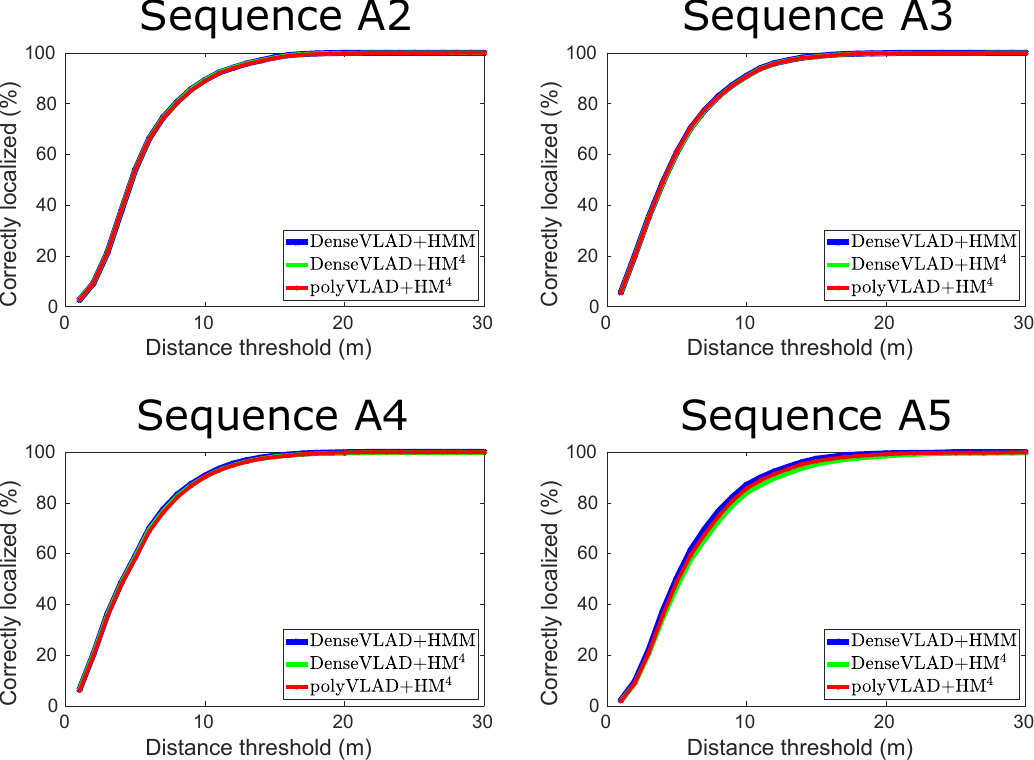}
		    \label{fig:ablation_accuracy}
		}
	}
	\mbox
	{
	    \subfloat[][]
		{
		    \includegraphics[width=0.44\textwidth]{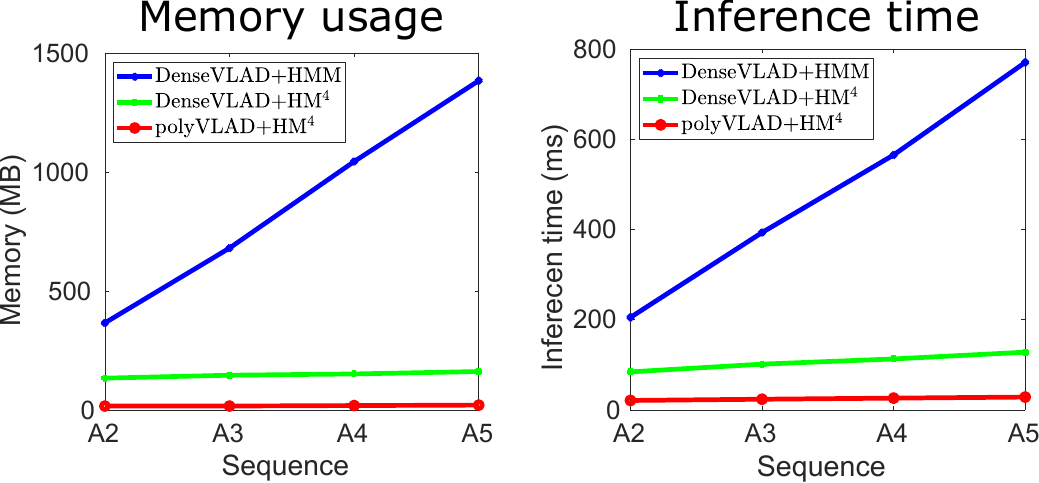}
		    \label{fig:ablation_scalability}
		}
	}
	
	\caption{Ablation study: \textbf{(a)} localization accuracy \textbf{(b)} growth in memory usage (in MB) \& inference time (millisecond/image)}
	\vspace{-0.6em}
\end{figure}

This experiment investigates the contribution of HM$^4$ and polyVLAD to the system. To this end, we use DenseVLAD incorporated with HM$^4$ (denoted as \texttt{DenseVLAD+HM$^4$}), which is then compared to \texttt{DenseVLAD+HMM} and \texttt{polyVLAD+HM$^{4}$}. Dataset St Lucia is used in this experiment

Fig.~\ref{fig:ablation_accuracy} demonstrates a comparable localization accuracy among these methods. More importantly, as shown in Fig.~\ref{fig:ablation_scalability}, regardless the image representation method, HM$^4$ offers the scalability in terms of both memory usage and inference time (i.e., sublinear growth), while HMM shows a linear growth. With the use of polyVLAD, the memory usage and inference time is significantly reduced by a constant factor. This experimental result is also consistent to the complexity analysis in section \ref{sec:complexity_analysis_system}.

In practice, if 1 sequence is collected every day to update the map, with only 5 sequences ($\sim$ operations in 5 days), the memory footprint \& inference time of HMM increase about $3\times$ and $4\times$ respectively. Hence, the algorithms with linear complexity prevent their applicability in embedded systems with a small hardware capability (e.g., self-driving cars), which also must concurrently process many other tasks. Therefore, optimizing the memory \& time complexity (as shown by \texttt{polyVLAD+HM$^{4}$}) in every task is crucial. 

\begin{figure*}[h]
	\centering
	\includegraphics[width=0.75\textwidth]{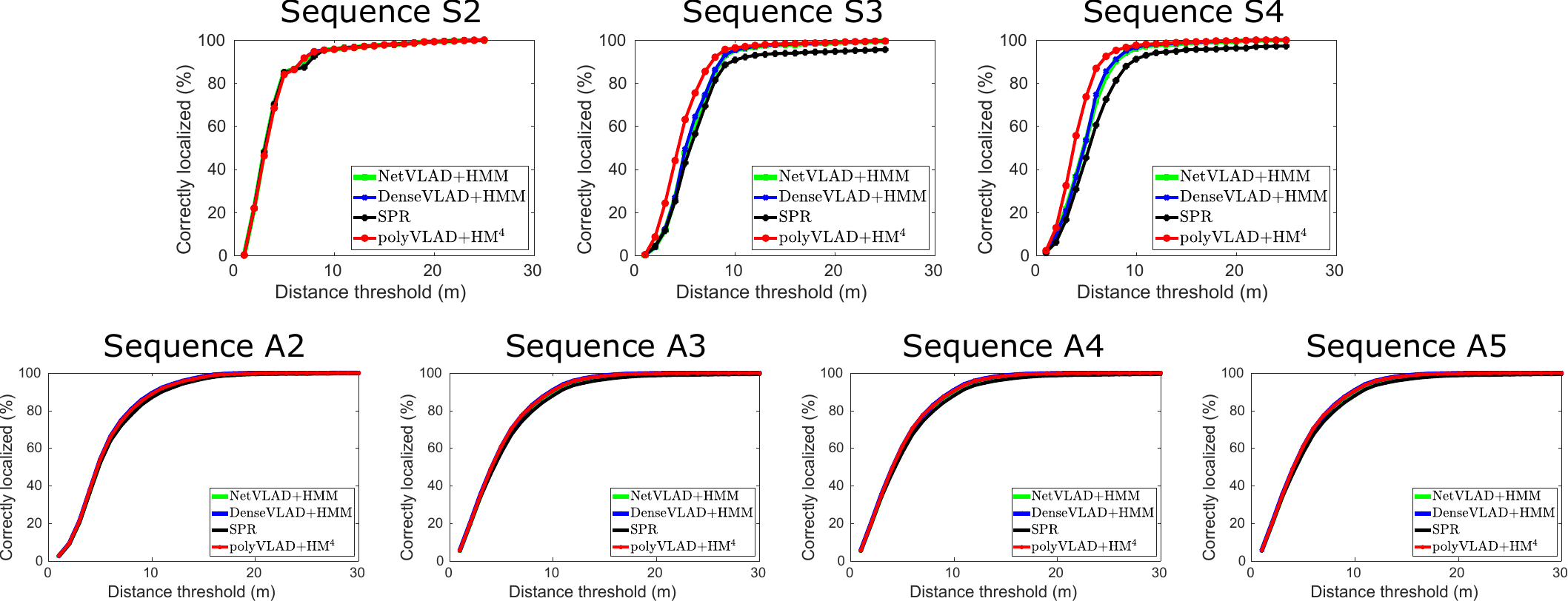}
	
	\caption{Localization accuracy on Oxford RobotCar (top) and St Lucia (bottom)}
	\label{fig:baseline_accuracy}
	\vspace{-1.2em}
\end{figure*}

\subsection{Comparison to baseline}

\begin{figure}
	\centering
	\mbox
	{
		\subfloat[][Growth in memory usage]
		{
			\includegraphics[width=0.4\textwidth]{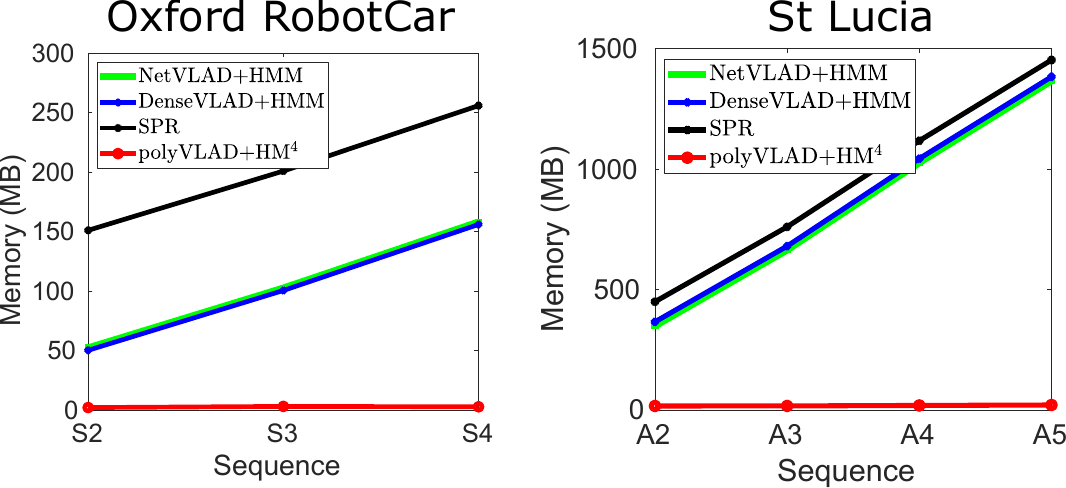}
			\label{fig:baseline_memory}
		}
	}
	\mbox
	{
		\subfloat[][Growth in inference time]
		{
			\includegraphics[width=0.4\textwidth]{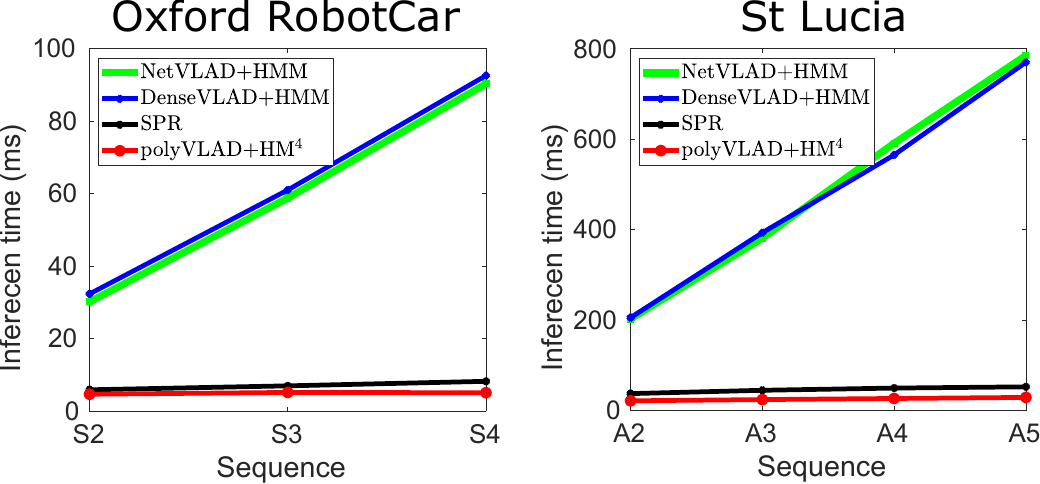}
			\label{fig:baseline_time}
		}	
	}
	
	\caption{The growth in \textbf{(a)} memory usage (in MB), and \textbf{(b)} inference time (millisecond/image).}
	\label{fig:baseline_scalability}
	\vspace{-0.5em}
\end{figure}

The growth in memory consumption and inference time is shown in Fig.~\ref{fig:baseline_scalability}, where the average value is reported, and the deviation is insignificant. The result is consistent in both Oxford RobotCar and St Lucia. Specifically, the memory usage and inference time of \texttt{NetVLAD+HMM} and \texttt{DenseVLAD+HMM} linearly grows when we update the new image sequences (Figs.~\ref{fig:baseline_memory} and \ref{fig:baseline_time}). For \texttt{SPR}, although its inference time is scalable (Fig.~\ref{fig:baseline_time}), its memory usage still linearly grows (Fig.~\ref{fig:baseline_memory}) (Note that \texttt{SPR} requires more memory than \texttt{NetVLAD+HMM} and \texttt{DenseVLAD+HMM} because it also stores K-means tree for retrieval). By contrast, \texttt{polyVLAD+HM$^4$} shows a sublinear growth in both memory usage and inference time (Fig.~\ref{fig:baseline_scalability})

 Regarding the localization accuracy (Fig.~\ref{fig:baseline_accuracy}), all methods share a comparable performance in general. \texttt{polyVLAD+HM$^4$} is slightly better than other methods in Oxford RobotCar.

\vspace{-0.5em}
\section{Conclusion} \label{sec:conclusion}
\vspace{-0.3em}

In reality, robots must concurrently process many tasks (e.g., segmentation, recognition, localization, etc). Due to the limitation of embedded hardware, it requires those tasks must have the scalable capability. In VPR, as robots continuously move and accumulate data, the memory needed for VPR increases unboundedly. HM$^4$ alleviates this problem by looking into the future and fetching just the right amount data needed to make VPR decisions. Combined with polyVLAD -- a lightweight and efficient image signature, we show the scalable capability of our method under the context of sequentially updating VPR system with new image sequences. In practice, our method can be deployed locally on an embedded hardware, in which the passive storage needs to be updated after few years. Another option is to implement our approach on a server-client architecture, in which the VPR task is shared between client and server. Also, this option leverages the nearly infinite storage of cloud and high-speed connection to transmit data between server and client.
\vspace{-0.7em}

\bibliographystyle{IEEEtran}
\bibliography{IEEEabrv,egbib}

\begin{thebibliography}{10}
\providecommand{\url}[1]{#1}
\csname url@rmstyle\endcsname
\providecommand{\newblock}{\relax}
\providecommand{\bibinfo}[2]{#2}
\providecommand\BIBentrySTDinterwordspacing{\spaceskip=0pt\relax}
\providecommand\BIBentryALTinterwordstretchfactor{4}
\providecommand\BIBentryALTinterwordspacing{\spaceskip=\fontdimen2\font plus
\BIBentryALTinterwordstretchfactor\fontdimen3\font minus
  \fontdimen4\font\relax}
\providecommand\BIBforeignlanguage[2]{{%
\expandafter\ifx\csname l@#1\endcsname\relax
\typeout{** WARNING: IEEEtran.bst: No hyphenation pattern has been}%
\typeout{** loaded for the language `#1'. Using the pattern for}%
\typeout{** the default language instead.}%
\else
\language=\csname l@#1\endcsname
\fi
#2}}

\bibitem{torii201524}
A.~Torii, R.~Arandjelovic, J.~Sivic, M.~Okutomi, and T.~Pajdla, ``24/7 place
  recognition by view synthesis,'' in \emph{CVPR}, 2015.

\bibitem{arandjelovic2016netvlad}
R.~Arandjelovic, P.~Gronat, A.~Torii, T.~Pajdla, and J.~Sivic, ``{NetVLAD}:
  {CNN} architecture for weakly supervised place recognition,'' in \emph{CVPR},
  2016.

\bibitem{churchill2013experience}
W.~Churchill and P.~Newman, ``Experience-based navigation for long-term
  localisation,'' \emph{IJRR}, 2013.

\bibitem{doan2019scalable}
A.-D. Doan, Y.~Latif, T.-J. Chin, Y.~Liu, T.-T. Do, and I.~Reid, ``Scalable
  place recognition under appearance change for autonomous driving,'' in
  \emph{ICCV}, 2019.

\bibitem{dymczyk2015gist}
M.~Dymczyk, S.~Lynen, T.~Cieslewski, M.~Bosse, R.~Siegwart, and P.~Furgale,
  ``The gist of maps-summarizing experience for lifelong localization,'' in
  \emph{ICRA}, 2015.

\bibitem{brahmbhatt2018geometry}
S.~Brahmbhatt, J.~Gu, K.~Kim, J.~Hays, and J.~Kautz, ``Geometry-aware learning
  of maps for camera localization,'' in \emph{CVPR}, 2018.

\bibitem{muja2014scalable}
M.~Muja and D.~G. Lowe, ``Scalable nearest neighbor algorithms for high
  dimensional data,'' \emph{TPAMI}, 2014.

\bibitem{ni2009epitomic}
K.~Ni, A.~Kannan, A.~Criminisi, and J.~Winn, ``Epitomic location recognition,''
  \emph{TPAMI}, 2009.

\bibitem{labbe2013appearance}
M.~Labbe and F.~Michaud, ``Appearance-based loop closure detection for online
  large-scale and long-term operation,'' \emph{T-RO}, 2013.

\bibitem{jegou2010aggregating}
H.~J{\'e}gou, M.~Douze, C.~Schmid, and P.~P{\'e}rez, ``Aggregating local
  descriptors into a compact image representation,'' in \emph{CVPR}, 2010.

\bibitem{andoni2015practical}
A.~Andoni, P.~Indyk, T.~Laarhoven, I.~Razenshteyn, and L.~Schmidt, ``Practical
  and optimal lsh for angular distance,'' in \emph{NIPS}, 2015.

\bibitem{cummins2008fab}
M.~Cummins and P.~Newman, ``Fab-map: Probabilistic localization and mapping in
  the space of appearance,'' \emph{IJRR}, 2008.

\bibitem{milford2012seqslam}
M.~J. Milford and G.~F. Wyeth, ``Seqslam: Visual route-based navigation for
  sunny summer days and stormy winter nights,'' in \emph{ICRA}, 2012.

\bibitem{latif2018addressing}
Y.~Latif, R.~Garg, M.~Milford, and I.~Reid, ``Addressing challenging place
  recognition tasks using generative adversarial networks,'' in \emph{ICRA},
  2018.

\bibitem{latif2020sprint}
Y.~Latif, A.-D. Doan, T.-J. Chin, and I.~Reid, ``Sprint: Subgraph place
  recognition for intelligent transportation,'' in \emph{ICRA}, 2020.

\bibitem{galvez2012bags}
D.~G{\'a}lvez-L{\'o}pez and J.~D. Tardos, ``Bags of binary words for fast place
  recognition in image sequences,'' \emph{T-RO}, 2012.

\bibitem{garg2020fast}
S.~Garg and M.~Milford, ``Fast, compact and highly scalable visual place
  recognition through sequence-based matching of overloaded representations,''
  \emph{ICRA}, 2020.

\bibitem{lowry2018lightweight}
S.~Lowry and H.~Andreasson, ``Lightweight, viewpoint-invariant visual place
  recognition in changing environments,'' \emph{RA-L}, 2018.

\bibitem{gong2014multi}
Y.~Gong, L.~Wang, R.~Guo, and S.~Lazebnik, ``Multi-scale orderless pooling of
  deep convolutional activation features,'' in \emph{ECCV}, 2014.

\bibitem{arandjelovic2013all}
R.~Arandjelovic and A.~Zisserman, ``All about vlad,'' in \emph{CVPR}, 2013.

\bibitem{kosub2019note}
S.~Kosub, ``A note on the triangle inequality for the jaccard distance,''
  \emph{Pattern Recognition Letters}, vol. 120, pp. 36--38, 2019.

\bibitem{huang1998extensions}
Z.~Huang, ``Extensions to the k-means algorithm for clustering large data sets
  with categorical values,'' \emph{Data mining and knowledge discovery}, 1998.

\bibitem{maddern20171}
W.~Maddern, G.~Pascoe, C.~Linegar, and P.~Newman, ``1 year, 1000 km: The oxford
  robotcar dataset,'' \emph{IJRR}, 2017.

\bibitem{glover2010fab}
A.~J. Glover, W.~P. Maddern, M.~J. Milford, and G.~F. Wyeth,
  ``{FAB-MAP}+{RatSLAM}: Appearance-based {SLAM} for multiple times of day,''
  in \emph{ICRA}, 2010.

\end{thebibliography}

\end{document}